%% file: samplepaper.tex
\documentclass[runningheads]{llncs}
\usepackage[T1]{fontenc}
\usepackage{graphicx}
\usepackage{microtype}
\usepackage{subfigure}

\usepackage{hyperref}
\usepackage{bbm}

\usepackage{booktabs}       
\usepackage{amsfonts}       
\usepackage{nicefrac}       
\usepackage{microtype}      
\usepackage{xcolor}         
\usepackage{tikz}
\usetikzlibrary{arrows}
\usetikzlibrary{positioning,chains,arrows}
\usetikzlibrary{decorations.pathmorphing}
\usetikzlibrary{decorations.markings}
\usetikzlibrary{decorations.pathreplacing}
\usetikzlibrary{calc}
\usetikzlibrary{arrows.meta}
\usetikzlibrary{bending}
\usetikzlibrary{intersections}
\usetikzlibrary{fadings}
\usetikzlibrary{babel}
\usetikzlibrary{shapes}
\definecolor{out-col}{RGB}{230,230,230}

\definecolor{H1-1-col}{RGB}{255,217,47}
\definecolor{H1-2-col}{RGB}{153,0,194}
\definecolor{H1-3-col}{RGB}{55,126,184}
\definecolor{H2-1-col}{RGB}{255,127,0}
\definecolor{H2-2-col}{RGB}{77,175,74}
\definecolor{H2-3-col}{RGB}{152,78,163}

\tikzstyle{tips}=[%
btip/.tip = {Latex[length=6, width=6]},
-btip]
\tikzstyle{neuralnetwork}=[%
draw=black,
node distance = 3mm and 18mm,
start chain = going right,
neuron/.style = {rectangle, draw=black,
	minimum size=6pt, inner sep=4pt,
	on chain},
annot/.style = {text width=4em, align=center,},
tips]

\usepackage[many]{tcolorbox}
\newtcolorbox{mybox}[2]{%
    tikznode boxed title,
    enhanced,
    colframe={#2},
    arc=0mm,
    boxrule=3pt,
    interior style={white},
    attach boxed title to top left= {yshift=-\tcboxedtitleheight/2, xshift=20mm},
    fonttitle=\bfseries,
    colbacktitle=white,coltitle={#2},
    boxed title style={size=normal,colframe=white,boxrule=0pt},
    title=\fontsize{30pt}{24}\selectfont{#1}}

\usepackage{tikz}
\usetikzlibrary{shapes}
\usetikzlibrary{decorations.pathreplacing}
\usetikzlibrary{arrows.meta}
\usetikzlibrary{calc}
\usetikzlibrary{shapes}

\makeatletter
\newdimen\@myBoxHeight%
\newdimen\@myBoxDepth%
\newdimen\@myBoxWidth%
\newdimen\@myBoxSize%
\newcommand{\SquareBox}[2][]{%
    \settoheight{\@myBoxHeight}{#2}
    \settodepth{\@myBoxDepth}{#2}
    \settowidth{\@myBoxWidth}{#2}
    \pgfmathsetlength{\@myBoxSize}{max(\@myBoxWidth,(\@myBoxHeight+\@myBoxDepth))}%
    \tikz \node [shape=rectangle, shape aspect=1,draw=red,inner sep=2\pgflinewidth, minimum size=\@myBoxSize,#1] {#2};%
}%
\makeatother

\tikzset{font={\fontsize{25pt}{24}\selectfont}}
\tikzset{operator/.style={rectangle, draw, inner sep=0pt, minimum size=2cm}}
\tikzset{mytip/.tip={>[length=12, width=16]}}

\definecolor{PaleBlue}{rgb}{0,.55,.9}
\definecolor{PaleGreen}{rgb}{0,.7,.25}
\definecolor{DarkGreen}{rgb}{0,.6,.25}
\definecolor{RedPink}{rgb}{.9,0,.2}
\definecolor{Pink}{rgb}{.85,.35,.7}
\definecolor{Purple}{rgb}{.6,0,.75}
\definecolor{Orange}{rgb}{.9,.3,.05}
\definecolor{GoldUL}{rgb}{1,.76,.32}

\colorlet{attentionColor}{Orange}
\colorlet{charEmbedColor}{RedPink}
\colorlet{predEmbedColor}{Pink}

\tikzstyle{embed}=[%
  draw,
  #1,
  anchor=north,
  minimum width=.8cm,
  minimum height=1.6cm,
  inner sep=0pt,
  text=#1!65!black,
  font=\fontsize{25pt}{24}\selectfont,
  ]

\usepackage{amsmath} 
\usepackage{mathtools,xparse}
\usepackage{graphicx}
\usepackage{xcolor}
\usepackage{algorithm}
\usepackage{algorithmic}
\usepackage{tikz,pgf}
\usetikzlibrary{matrix,shapes,arrows,positioning}
\usepackage{enumitem}
\usepackage{color}
\usepackage{colortbl}
\definecolor{Gray}{gray}{0.9}

\newtheorem{prop}{Proposition}

\newtheorem{improvement}{Improvement}

\def\equationautorefname~#1\null{%
  Equation~(#1)\null
}

\usepackage[toc,page,header]{appendix}
\usepackage{minitoc}
\renewcommand \thepart{}
\renewcommand \partname{}

\usepackage{listings}

\usepackage{multirow}
\usepackage{definitions}

\newcommand{\citep}{\cite}
\newcommand{\citet}{\cite}

\begin{document}
\title{Seeking Interpretability and Explainability in Binary Activated Neural Networks}
\titlerunning{Seeking Interpretability and Explainability in Binary Activated NNs}
%
%
\author{Benjamin Leblanc\orcidID{0009-0006-0971-3463} and
Pascal Germain\orcidID{0000-0003-3998-9533}}
\authorrunning{Leblanc and Germain}
%
\institute{Université Laval, Québec, Canada \\
\email{benjamin.leblanc.2@ulaval.ca} \qquad
\email{pascal.germain@ift.ulaval.ca}}
\maketitle              
\begin{abstract}
We study the use of binary activated neural networks as interpretable and explainable predictors in the context of regression tasks on tabular data. We first dissect those specific networks to understand their inner workings better and bound their best achievable training performances. We then use this analysis as a theoretical foundation to propose a greedy algorithm for building interpretable binary activated networks. The simplicity of the predictor being instrumental for achieving interpretability, our approach builds the predictors one neuron at a time, so that their architecture (complexity) suits the task at hand. Finally, we present an approach based on the efficient computation of SHAP values for quantifying the relative importance of the features, hidden neurons and individual connections (weights) of these particular networks. Our work sets forth a new family of predictors to consider when interpretability is of importance.

\end{abstract}

\section{Introduction}
\label{section:introduction}
\input{sections/introduction}

\section{Related Works and Positioning}
\label{section:related}
\input{sections/related}

\section{Notation}
\label{section:background}
\input{sections/background}

\section{Dissecting Binary Activated Neural Networks}
\label{section:BANN}
\input{sections/BANN}

\section{The BGN (Binary Greedy Network) Algorithm}
\label{section:algorithm}
\input{sections/algorithm}

\section{SHAP values for BANNs: inputs, neurons and weights}
\label{section:BANN SHAP}
\input{sections/BANN_SHAP}

\section{Numerical Experiments}
\label{section:experiments}
\input{sections/experiments}

\section{Conclusion}
\label{section:conclusion}
\input{sections/conclusion}

\subsection*{Acknowledgements}
This research was funded by the NSERC/Intact Financial Corporation Industrial Research Chair in Machine Learning for Insurance. Pascal Germain is supported by the Canada CIFAR AI Chair Program, and the NSERC Discovery grant RGPIN-2020-07223.

\subsection*{Disclosure of Interests}

The authors have no competing interests relative to the content of this article. 

\bibliographystyle{splncs04}
\bibliography{references}

\appendix

\section{Supplementary material}
\label{section:sup}
\input{sections/supplementary}

\section{Mathematical results}
\label{section:math}
\input{sections/math}

\end{document}

%% file: sections/introduction.tex
Among machine learning models, rule-based predictors tend to be recommended on tasks where the interpretability of the decision process is of importance. Notably, decision trees are a common choice 
\cite{molnar2022,DBLP:journals/corr/abs-2103-11251}, for they can be efficient predictors - accurate, yet quite simple. This contrasts with neural networks (NNs), generally perceived as opaque black boxes. Various techniques have been proposed to lighten deep neural networks (DNNs), making them easier to interpret: in addition to being a natural regularizer \citep{DBLP:conf/nips/HubaraCSEB16,DBLP:conf/iclr/LinGH19}, the use of weights encoded by a small number of bits \citep{DBLP:conf/iclr/AlizadehFLG19,DBLP:conf/nips/CourbariauxBD15,DBLP:conf/icml/MengBK20,DBLP:journals/pr/QinGLBSS20} and quantized activation functions, notably binary ones \citep{DBLP:conf/nips/SoudryHM14,DBLP:conf/nips/HubaraCSEB16,DBLP:conf/cvpr/WangLT0019}, have shown promising results. Also, methods such as network pruning \citep{DBLP:conf/nips/HanPTD15,DBLP:conf/iclr/RendaFC20,DBLP:journals/corr/HuPTT16,DBLP:journals/corr/IandolaMAHDK16,DBLP:conf/mobisys/GoetschalckxMWV18}, weight sharing \citep{DBLP:conf/icml/ChenWTWC15} or matrix factorization \citep{DBLP:conf/bmvc/JaderbergVZ14} allow compacting a trained NN, resulting in predictors having fewer parameters.

In this work, we tackle the following question: could neural networks be competitive to models such as tree-based approaches when both empirical performances and interpretability are critical? To do so, we focus on binary activated neural networks (BANNs), i.e.\ NNs where all of the activation functions are binary steps; such a constraint on the models fosters interpretability, as they express thresholds. We argue that for several learning tasks (here, regression tasks on tabular data), BANNs with a \emph{light architecture} can be built to both obtain competitive performances and be interpretable to human beings, even though it is well-known that tree-based approaches outperform complex neural network approaches on such tasks \cite{DBLP:conf/nips/GrinsztajnOV22}.

We first derive mathematical observations that help understand the inner workings of the BANNs family of predictors. This analysis provides the theoretical underpinnings of our second contribution: a greedy algorithm denominated BGN (Binary Greedy Networks) for building interpretable BANNs. Namely, by iteratively building the network one neuron at a time, BGN leads to predictors whose complexity is tailored to the task at hand. Thirdly, we introduce a method, based on SHAP values \cite{DBLP:conf/nips/LundbergL17}, for efficiently computing complementary explanations about BANNs behavior. The user can assess the relative importance of each input feature in the decision process (as typically provided by SHAP values), but also the importance of every neuron and their interconnections. Finally, we present comprehensive empirical experiments, showing BGN effectively yields efficient predictors, competitive to other BANN training algorithms when it comes to predictive accuracy, and regression trees when it comes to interpretability. Globally, our work sets forth a new family of predictors to consider when interpretability is of importance.

%% file: sections/related.tex
\paragraph{Interpretability and explainability.}
In line with Rudin \cite{DBLP:journals/natmi/Rudin19}, we consider that the degree of \textit{interpretability} of a predictor should be judged by the capacity of a non-expert to understand its decision process solely by considering the predictor in itself (this notion is also referred to as \textit{ante-hoc explainability}). In this context, binary activation functions acting as thresholds (just as in regression trees) is arguably better suited than ReLU or tanh activations for interpretation purposes. We focus on the following aspects of BANNs (among others \cite{DBLP:journals/corr/abs-2103-11251}) impacting their interpretability: the number of features it uses (\textit{parsimony}), its width, depth, and its number of non-zero weights (\textit{sparseness}). We call \textit{efficient} (or \textit{compact}) a predictor achieving a good tradeoff between these aspects and predictive performances.

This view on interpretability greatly differs from \textit{explaining} a model (\textit{post-hoc explainability}), where explanations present inherently hidden information on the model concerning its decision process. For example, SHAP values \cite{DBLP:conf/nips/LundbergL17} explain the impact of the different features of a model on its predictions. These explanations, in machine learning, are usually simplifications of the original model \cite{DBLP:journals/corr/abs-2009-04521,DBLP:conf/aaai/SubramanianPJBH18,DBLP:conf/kdd/Ribeiro0G16}.

There are a few machine learning models to be considered state-of-the-art regarding interpretability \cite{DBLP:journals/corr/abs-2103-11251}: decision trees, decision lists (rule lists), scoring systems, linear models, etc. Even though efforts have been put into explaining neural networks, up to the author's knowledge, they have never been considered interpretable predictors for regression purposes.
In this work, in addition to promoting BANNs interpretability, we suggest that additional insights on such models can be obtained by explaining the decision process. 

\paragraph{Binary activated neural networks (BANNs).} 
Neural network learning mostly relies on stochastic gradient descent (SGD), which requires the objective function to be fully differentiable. Because the derivative of step functions is zero almost everywhere, this technique cannot directly be used for training with such activation functions. Most of the literature concerning BANNs proposes workarounds allowing the use of the SGD algorithm. A simple way to estimate the gradient of the binary function is the straight-through estimator \cite{DBLP:journals/corr/BengioLC13,DBLP:conf/nips/HubaraCSEB16}, which uses the identity function as a surrogate of the gradient and which has then been refined over time \cite{DBLP:journals/corr/abs-1812-11800,DBLP:conf/eccv/LiuWLYLC18}. While such methods are convenient, they lack solid theoretical groundings. It has also been proposed to use continuous binarization \cite{DBLP:conf/iccv/GongLJLHLYY19,DBLP:conf/cvpr/YangSXTLDH019,DBLP:conf/icassp/SakrCWGS18}; continuous activation functions are used, increasingly resembling a binary activation functions over the training iterations. Another branch of BANNs training algorithms consists in assuming a probability distribution on its weights; doing so, one can work with the expectation of each layer output and train BANNs with the SGD algorithm \cite{DBLP:conf/nips/SoudryHM14,DBLP:conf/nips/LetarteGGL19}. 

\paragraph{Pruning.} 
When one seeks a \emph{light} (interpretable) network, a common approach is to train \emph{heavy} networks at first and then to prune it \cite{DBLP:journals/corr/abs-1903-01611,DBLP:journals/corr/abs-1902-09574,DBLP:conf/nips/HanPTD15,DBLP:journals/corr/HanMD15}. As implied by the Lottery Ticket Hypothesis \cite{DBLP:conf/iclr/FrankleC19}, for the approach to work well, the heavy network must be big enough so that the chances are high that it contains an efficient (compact but performing) sub-network. Working backward would be an interesting avenue for obtaining such networks without needing important computational resources to train wide and deep NNs first; starting with small NNs and making them grow, just like a decision tree is built. A convenient way to do so is by using a greedy approach.

\paragraph{Greedy neural network training.} Greedy approaches for training DNNs usually refers to the use of a layer-wise training scheme \cite{DBLP:journals/corr/KulkarniK17,DBLP:conf/nips/BengioLPL06,DBLP:journals/corr/abs-2107-04466,DBLP:conf/evoW/CustodeTBCCV20,DBLP:journals/corr/abs-1905-10409}. The usual procedure is the following: first, the full network is trained. Then, the first layer of the network is fixed. The network is retrained, where each layer but the first is fine-tuned. The second layer of the network is then fixed, and so on.
 While it has been shown that greedy approaches can scale to large datasets such as ImageNet \citep{DBLP:conf/icml/BelilovskyEO19}, the literature concerning greedy approaches for training NNs is scarce and does not exploit the idea that such a method could lead to interpretable predictors.

%% file: sections/background.tex
Our study is devoted to univariate regression tasks on tabular data. Each task is characterized by a dataset $S \,{=}\, \{(\mathbf{x}_i,y_i)\}_{i=1}^m$ containing $m$ instances, each one described by features $\mathbf{x}\in\mathcal{X}\subseteq \mathbb{R}^d$ and labels $y\in\Rbb$. From this dataset, we aim to train a binary activated neural network (BANN), {\it i.e.}, with activation functions on hidden layers having two possible output values. From now on, we will focus on the sign activation function:
$$
\sgn(x) = \begin{aligned}
    	\begin{dcases}
    	-1&\textup{if } x < 0,\\
    	+1&\textup{otherwise}.
    	\end{dcases}
    	\end{aligned}
$$
We consider fully-connected BANNs composed of $l \in \mathbb{N}^*$ layers $L_k$ of size (width) $d_k$, for $ k \in \{1,\dots,l\}$. We fix the output layer size to $d_l = 1$, and denote $d_0 = d$ the input dimension.
We call $l$ the depth of the network and $d^* = \max_{k \in \{1, 2, \dots, l-1\}}{d_k}$ the width of the network. The sequence $\mathbf{d}\eqdots\langle d_k\rangle_{k=0}^{l}$ encapsulates the neural network architecture. We denote a BANN predictive function by $B:\Rbb^d\to\Rbb$. Every predictor $B$ is characterized by the values of its weights $\{\Wbf_k\}_{k=1}^{l}$ and biases $\{\bbf_k\}_{k=1}^{l}$, where $\Wbf_k \in \mathbb{R}^{d_{k-1} \times d_k}$ and $\bbf_k \in \mathbb{R}^{d_k}$. Each layer $L_k$ is a function 
\begin{equation}\label{eq:layer_function}
L_k(\xbf) = f_k\left(\Wbf_k\xbf+\bbf_k\right) ,
\end{equation}
where $f_k(\cdot)$, $k\in\{1,\dots,l-1\}$, are binary activation functions acting element-wise; the output layer's activation function $f_l(\cdot)$ being a linear activation (identity) function, allowing predicting real-valued output. For $i,j \in \{1,\dots,l\}$ and $ i<j$, we denote 
$$L_{i:j}(\xbf) = (L_j \circ L_{j-1} \circ \dots \circ L_i)(\xbf).$$ The whole predictor $B$ is given by the composition of all of its layers: $B(\xbf) {=} L_{1:l}(\xbf)$. While $\mathcal{X}\subseteq\Rbb^d$ denotes the input space (domain) of the network, $\mathcal{L}_{i:j}$ denotes the image of the function given by the composition of layers $L_i$ to $L_j$ (with $\mathcal{L}_{i}\eqdots\mathcal{L}_{i:i}$) and $\mathcal{B}\eqdots \Lcal_{1:l}$ the image of the whole network.

%% file: sections/BANN.tex
\begin{figure}[t]
\begin{center}
\includegraphics[width=0.7\linewidth]{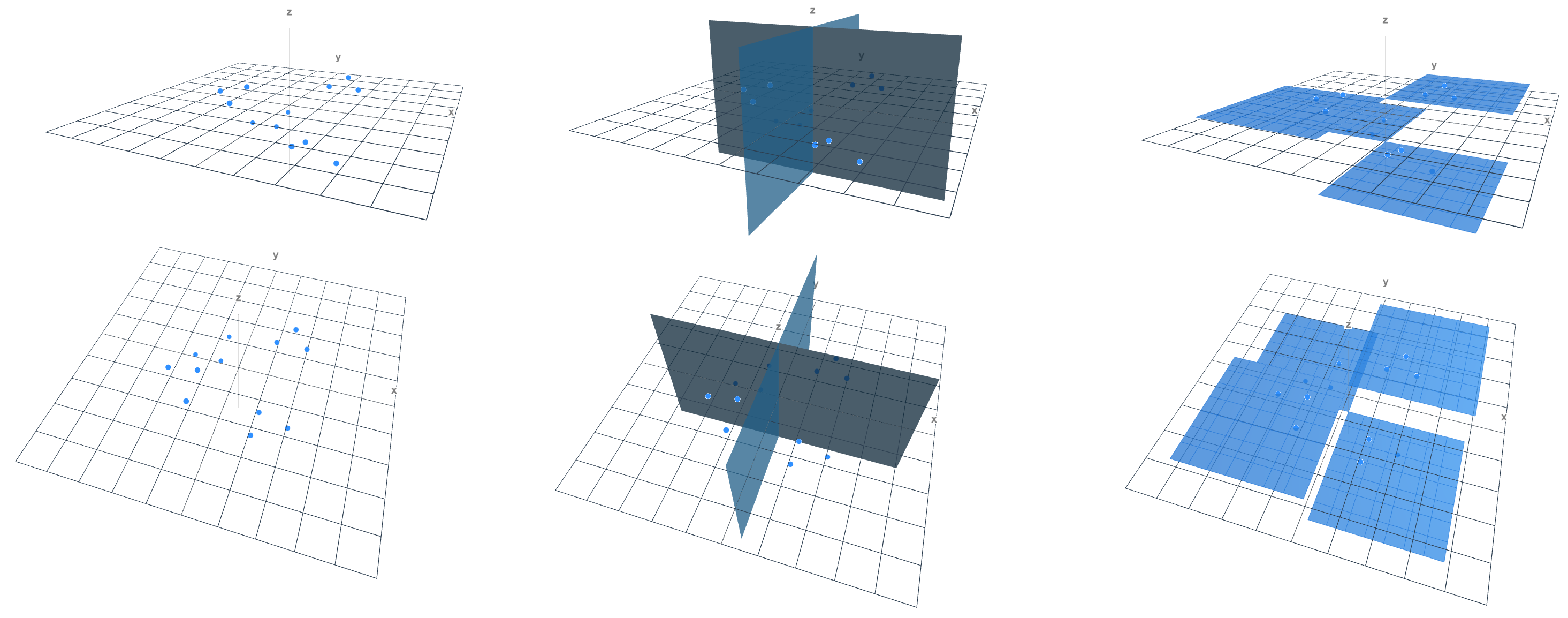} \caption{The behavior of the different layers of a single-hidden layer BANN with architecture $\mathbf{d} = \langle 2,2,1\rangle$ acting in the input space $\mathcal{X}$. The upper figures correspond to their lower counterparts, with a different point of view. Left: The x-axis and y-axis represent the feature space of a 2D problem, the z-axis being the label space. Middle: the hidden layer of the network separates the input space into four regions with the help of $d_1 = 2$ hyperplanes. Right: predictions rendered by the network as a function of its input; the predictions are constant within each region defined by the hidden layer.} \label{fig:hyperplanes}
\vspace{-4mm}
\end{center}
\end{figure}

Each neuron of the first hidden layer ($L_1$) of a BANN can be seen as a hyperplane acting in its input space $\mathcal{X}$. Indeed, given an input $\xbf\in\Xcal$, these neurons apply a binary threshold on a linear transformation of $\xbf$ (see \autoref{eq:layer_function}). If $d_1=1$, then $L_1(\mathbf{x})$ indicates whether $\xbf$ is on one side ($+1$) or the other ($-1$) of a hyperplane parameterized by the coefficients $\{\wbf_1,b_1\}$ characterizing that first hidden layer. More generally, with $d_1 \in \mathbb{N}^*$, $L_1$ is a hyperplane arrangement whose output $L_1(\xbf) \in \{+1,-1\}^{d_1}$ indicates to which region a specific $\xbf$ belongs. \autoref{fig:hyperplanes} illustrates that phenomenon.

Furthermore, when a BANN has more than one hidden layer ($l > 2)$, the function \mbox{$L_{2:l-1}~:~\{-1,+1\}^{d_1} \rightarrow \{-1,+1\}^{d_{l-1}}$} groups together some regions established by the first layer $L_1$, so that inputs falling in different regions can share a same prediction. That is, for any \mbox{$\mathbf{x}, \mathbf{x}' \in \mathcal{X}$} and $k \in \{1,\dots,l-1\}$, we have that \mbox{$L_{1:k}(\mathbf{x}) = L_{1:k}(\mathbf{x}') \Rightarrow L_{1:k+1}(\mathbf{x}) = L_{1:k+1}(\mathbf{x}')$}. This property leads to the following relationship, which actually holds for any deterministic network: \mbox{$|\mathcal{L}_1| \geq |\mathcal{L}_{1:2}| \geq \dots \geq |\mathcal{L}_{1:l}|$}. Put into words, the image of a given layer $L_k$ cannot contain more elements than its domain, which is the image of the preceding layer~$L_{k-1}$.

These observations, combined with the fact that the predictions given by a BANN are constant within a given region, lead us to the following theorem\footnote{The proof of \autoref{prop:bin_bnd} is given in the appendices (\autoref{section:math}).}, establishing bounds on the training performance of the network in terms of train mean squared error (MSE): $\ell_S(B) = \frac1m \sum_{i=1}^{m} (B(\xbf_i)-y_i)^2$.

\begin{theorem}\label{prop:bin_bnd}
Let $B$ be a BANN of depth $l$ and $S$ a dataset of size $m$. We have 
$$
\ell_S(B) 
\geq\sum_{\mathclap{\pbf\in\mathcal{L}_{1:l}}}\frac{\left|\ybf_{\pbf}^{(l)}\right|}{m}\textup{Var}\left(\ybf_{\pbf}^{(l)}\right) \geq \dots \geq  \sum_{\mathclap{\pbf\in\mathcal{L}_{1:2}}}\frac{\left|\ybf_{\pbf}^{(2)}\right|}{m}\textup{Var}\left(\ybf_{\pbf}^{(2)}\right) \geq  \sum_{\mathclap{\pbf\in\mathcal{L}_1}}\frac{\left|\ybf_{\pbf}^{(1)}\right|}{m}\textup{Var}\left(\ybf_{\pbf}^{(1)}\right),
$$
where $\ybf_{\pbf}^{(k)} = \{y\ |\  (\mathbf{x},y)\in S \land (L_k \circ \dots \circ L_1)(\xbf) = \pbf\} \ \forall k\in\{1,\dots,l\}$.
\end{theorem}

Note that the terms bounding the train MSE is the weighted sum of the label's variances within the various regions created by $L_1$ and tied together by a further layer. The proof relies on the fact that BANNs can't do better than predicting the average label value in each of these regions independently. 

\autoref{prop:bin_bnd} highlights the importance of the leading layer: a BANN with a poor division of $\mathcal{X}$ by $L_1$ is bound to yield poor predictive performances, no matter how the following layers are tuned. It also justifies the layer-wise approach: the most restricting bounding term only depends on $L_1$, while the second most restricting depends on both $L_1$ and $L_2$, and so on. 

%% file: sections/algorithm.tex
In the following, we propose our BGN learning algorithm. We emphasize that our design choices are driven by the intention to build an interpretable predictor. As mentioned in \autoref{section:related}, instead of pruning a neural network in a post-processing fashion, our objective is to directly train the network to be as light --- i.e., parsimonious, sparse, shallow, and narrow --- as possible.

We first introduce the core principles of our algorithm in the case of single-hidden layer BANNs before aiming toward deep architecture.

\subsection{Learning shallow networks} 

\begin{figure}[t]
    \centering
    \includegraphics[width=0.8\linewidth]{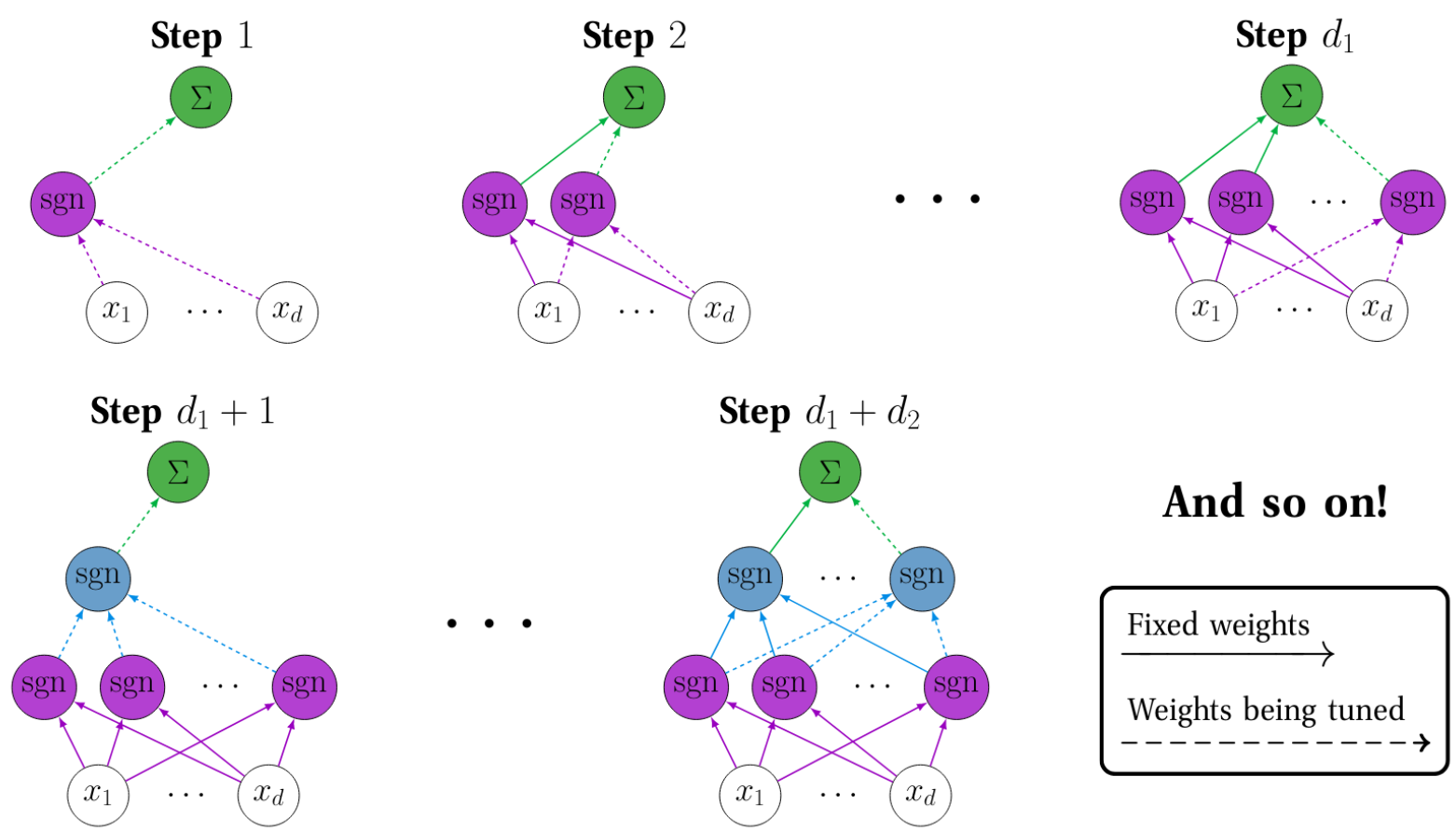}
    \caption{Building binary activated neural networks with the BGN algorithm, one neuron at a time, one layer at a time. Only weights are displayed, yet biases are tuned as well.}
    \label{fig:bgn_building}
\end{figure}

We start by considering single-hidden layer BANNs (that is, $l=2$). In order to learn a compact BANN architecture, the proposed BGN algorithm first considers a hidden layer of width $1$ and iteratively makes it grow by adding one neuron at a time. Thus, we start by building a network with architecture $\mathbf{d}^{(1)} = \langle d,1,d_2\rangle$, then adding a neuron to obtain an architecture $\mathbf{d}^{(2)} = \langle d,2,d_2\rangle$, and so on. The first $d_1$ steps from \autoref{fig:bgn_building} (upper part) illustrate the process. Each time a neuron is added, the bound 
$\sum_{\pbf\in\mathcal{L}_1}\frac{\left|\ybf_{\pbf,1}\right|}{m}\textup{Var}\left(\ybf_{\pbf,1}\right)$ from \autoref{prop:bin_bnd} is minimized so that the final predictor is guided by the lower bound on the obtainable MSE. 
That is, we aim to parameterize the first neuron of the hidden layer as 
\begin{equation}\label{eq:simple_hyperplane}
(\mathbf{w}_{1,1},b_{1,1}) \approx 
\underset{\mathbf{w},b}{\argmin}\ \left( \frac{|\ybf_{-1}|}{m}\textup{Var}\left(\ybf_{-1}\right)+\frac{|\ybf_{+1}|}{m}\textup{Var}\left(\ybf_{+1}\right)\right),
\end{equation}
with 
\begin{equation}\label{eq:ypm}
\ybf_{\pm1} = \{y_{i}\ |\ (\mathbf{x}_i,y_i)\in S,\ \sgn(\mathbf{w}\cdot\mathbf{x}_i+b) = \pm1\}.
\end{equation}
\begin{algorithm}[t]
	\caption{Binary Greedy Network (BGN) - One hidden layer BANN}\label{algo:bgn}
	\begin{algorithmic}[1]
	\STATE \textbf{Input} : $S = \{(\xbf_1,y_1),\dots,(\xbf_m,y_m)\}$, $y \in \mathbb{R}$, $\xbf\in\mathbb{R}^d$, a dataset
	\STATE Set $\mathbf{r}^{(1)} = \ybf = (y_1,\ldots, y_m)$, $t=0$ and $b_2 = 0$
	\STATE \textbf{While} stopping criterion is not met:
    \STATE $\qquad$ $t = t + 1$
 	\STATE $\qquad$ $\wbf_{1,t} = \textup{LassoRegression}(S^{(t)})$,
	\quad where $S^{(t)}= \{(\xbf_1,r_1^{(t)}),\dots,(\xbf_m,r_m^{(t)})\}$\label{line:heuristic}
    \STATE $\qquad$ $b_{1,t} = {\argmin}_b\ \left( \frac{|\mathbf{r}^{(t)}_{-1}|}{m}\textup{Var}\left(\mathbf{r}^{(t)}_{-1}\right)+\frac{|\mathbf{r}^{(t)}_{+1}|}{m}\textup{Var}\left(\mathbf{r}^{(t)}_{+1}\right)\right)$
    with $\mathbf{r}^{(t)}_{\pm1}$ given by Eq.~\eqref{eq:ypm}
	\STATE  $\qquad$ $\rho_\pm^{(t)} = \dfrac{\sum_{i:\sgn(\xbf_i\cdot\mathbf{w}_{1,t}+b_{1,t})=\pm1} r_i^{(t)}}{\sum_{i:\sgn(\xbf_i\cdot\mathbf{w}_{1,t}+b_{1,t})=\pm1} 1}$
    \STATE $\qquad$ $w_{2,t} = \frac{1}{2}\left(\rho_+^{(t)}-\rho_-^{(t)}\right); \quad  b_{2,t} = \frac{1}{2}\left(\rho_+^{(t)}+\rho_-^{(t)}\right)$ 
    \STATE $\qquad$ \textbf{For} $i \in \{1,\dots,m\}$:
    \STATE $\qquad\qquad$ $r^{(t+1)}_i=r^{(t)}_i - w_{2,t}\ \sgn(\mathbf{w}_{1,t}\cdot\mathbf{x}_i+b_{1,t}) - b_{2,t}$ 
    \STATE $\qquad$ $b_{2} = b_{2} + b_{2,t}$
	\STATE \textbf{Output} : $BGN_t(\xbf)$ = $\mathbf{w}_{2}\cdot\sgn(\Wbf_{1}\xbf+\mathbf{b}_{1})+b_{2}$\\
	\end{algorithmic} 
\end{algorithm}%
This amounts to positioning the hyperplane such that the examples with ``small" label values are separated from the examples with ``large" label values. However, since \autoref{eq:simple_hyperplane} is not a convex problem, we propose to use a linear regression algorithm as a proxy to obtain a weight vector $\mathbf{w}_{1,1}$ that points in the direction of the input space where the label values increase the most. More precisely, we use the Lasso regression \citep{tibshirani96} algorithm that also enforces the sparsity of the weight vector, and thus favors the interpretability of the model. 
Note that the bias parameter given by the linear regression algorithm is discarded as it would not translate into the $b$ value of \autoref{eq:ypm}that properly splits the labels into the $\ybf_{\pm1}$ sets.
Instead, finding $b$ minimizing \autoref{eq:simple_hyperplane} is done by sorting the examples of $S$ by their distance to the hyperplane parametrized by $\mathbf{w}_{1,1}$, and then computing the MSE associated with the $m+1$ possible splits of the sorted dataset. The parameter $b_{1,1}$ is then the one giving the largest separation margin between the sets $\ybf_{\pm1}$.

Once the first hidden neuron $(\mathbf{w}_{1,1},b_{1,1}) $ is fixed, the BGN algorithm computes the connecting weight $w_{2,1}$ of the output regression layer, and the bias term \mbox{$b_{2} = b_{2,1}$}. Then, the residual of the label values are computed, in a way that an example's label becomes the error of the predictor on that example. These residuals act as the new label values for finding the weights and bias of the second neuron $(\mathbf{w}_{1,2},b_{1,2})$ during the next iteration, as well as the weight $w_{2,2}$ of new the connection of the output layer and the updated output bias \mbox{$b_2 = b_{2,1} + b_{2,2}$}.\footnote{This is reminiscent of the Boosting algorithm \citep{DBLP:conf/icml/FreundS96}, where each new \emph{weak} predictor is added to compensate the error of the current \emph{ensemble}. In our setup, each \textit{weak} predictor is a hidden neuron.} The next subsection shows that the choices of $w_{2,t}$ and output bias correction term $b_{2,t}$ given by \autoref{algo:bgn} guarantee to minimize the training MSE at every iteration. Finally, note that the greedy process is repeated until a given stopping criteria is met; for the experiments of \autoref{section:experiments}, we consider the validation MSE loss not to decay for 100 epochs as the stopping criteria.

\subsection{Properties of the BGN algorithm}

Each time a neuron is added to the hidden layer, the weight of the connection to the output layer is computed as well as the correction to the output bias. Their value is such that it minimizes the train MSE of the network, as expressed by the next proposition.
\begin{prop}\label{prop:BGN_min}
In \autoref{algo:bgn}, we have
$$(w_{2,t}, b_{2,t}) = \underset{w,b}{\argmin}\sum_{i=1}^m \left(r_i^{(t)}-w\ \sgn(\mathbf{w}_{1,t}\cdot\xbf_i+b_{1,t})-b\right)^2.$$
\end{prop}
Furthermore, the objective we chose for our hyperplanes to minimize (\autoref{eq:simple_hyperplane}) not only optimizes the minimum obtainable MSE of the predictor but ensures that at every iteration of the algorithm, \textit{i.e}., every time a neuron is added to the network, the MSE obtained by the network on its training dataset diminishes. 
\begin{prop}\label{prop:BGN_dec}
Let $\ell_S(B)$ be the MSE of predictor $B$ on dataset $S$. Then,
$$
\ell_S(BGN_{t-1}) - \ell_S(BGN_{t}) = 
(w_{2,t}+b_{2,t})(w_{2,t}-b_{2,t}) > 0,~\forall t \in \mathbb{N}^*.
$$
\end{prop}
The proof of the two propositions above is given in the appendices (\autoref{section:math}).

\subsection{Improvements and deeper networks}

A key motivation for BGN is to find predictors with the right complexity for the task at hand. The greedy nature of the approach makes it such that a given neuron might become irrelevant as more neurons are added, which becomes a hurdle to obtaining a compact model. Thus, we experimentally assess that reconsidering the parameterization of the hidden neurons throughout the training yields to more compact predictors. 

\begin{improvement}[neuron replacement]\label{improv:1}
After performing an iteration $t>1$, a random hyperplane (i.e., a hidden neuron) is removed before doing another iteration of the algorithm. If the training error lowers, the new hyperplane (neuron) is kept; otherwise, the former is placed back. This is done $t$ times at each iteration.
\end{improvement}

The following improvement extends our BGN algorithm to deep BANNs, that is networks with two or more hidden layers. The resulting greedy procedure is illustrated by \autoref{fig:bgn_building} (lower part).

\begin{improvement}[adding hidden layers]\label{improv:2}
When the stopping criterion is met while building of the first hidden layer, that layer is fixed and a second hidden layer is built. The same iterative BGN algorithm is then applied, but the outputs of the first hidden layer now act as the input features to grow the second layer. This can be repeated for an arbitrary number of hidden layers; the BANN with the best overall validation MSE is kept.
\end{improvement}

The pseudo-code of the BGN algorithm with both \autoref{improv:1} and \autoref{improv:2} is presented in the appendices, \autoref{sec:bgn_algos}.

%% file: sections/BANN_SHAP.tex
In this section, we propose to complement the interpretability of BGN networks with information only explainability could provide: the relative importance of features, neurons, and weights in BANNs. To do so, we leverage a standard, yet most of the time intractable approach for computing SHAP values and adapt it to the particularities of simple BANNs to obtain tractable computation time. 

SHAP values \cite{DBLP:conf/nips/LundbergL17} is a metric quantifying the impact (magnitude) of a given feature value on the prediction of a given model for a given example. These latter are widely used due to their simplicity and theoretical groundings, but their computational cost, which is exponential in $d$ (the amount of input features), is truly limiting. Given a set of feature $D = \{1,\dots,d\}$, a single feature $i\in D$, an input $\mathbf{x}\in\Xcal$ and a predictor $f:\Rbb^d\to\Rbb$, the SHAP value is given by 
\begin{equation}\label{eq:shap_1}
\phi_i(f, \mathbf{x}) = \underbrace{\sum_{Q\subseteq D\setminus \{i\}}}_{\scriptsize\boxed{1}} \underbrace{\frac{|Q|!(|D|-|Q|-1)!}{|D|!}}_{\scriptsize\boxed{2}}\underbrace{\left(\mathbb{E}_{X_{\overline{Q\cup \{i\}}}}[f(\mathbf{x}_D)]-\mathbb{E}_{X_{\overline{Q}}}[f(\mathbf{x}_D)]\right)}_{\boxed{3}},
\end{equation}
where $\overline{Q} = D \setminus Q$ and $\mathbb{E}_{X_{\overline{Q}}}[f(\mathbf{x}_D)]$ is the expectation of the predictor output when the input features $Q$ of $\xbf$ are fixed but the other features $\overline{Q}$ are sampled according to the (empirical) data distribution. As explained in \autoref{section:algorithm}, a single-hidden layer BANN (\mbox{1-BANN}) can be seen as an aggregation of independent predictors (hidden neurons); we make use of this property of BANNs, and the additivity of the SHAP values to rewrite \autoref{eq:shap_1} for 1-BANNs:
$$
\phi_i(B, \mathbf{x}) = \hspace{-5mm}\underbrace{\sum_{k\in\{1,\dots,d_1\}}\hspace{-5mm}w_{2,k}}_{\scriptsize\boxed{0}}\underbrace{\sum_{Q\subseteq D\setminus \{i\}}}_{\scriptsize\boxed{1}} \underbrace{\tfrac{|Q|!(|D|-|Q|-1)!}{|D|!}}_{\scriptsize\boxed{2}}\underbrace{\left(\mathbb{E}_{X_{\overline{Q\cup \{i\}}}}[L_{1,k}(\mathbf{x}_D)]-\mathbb{E}_{X_{\overline{Q}}}[L_{1,k}(\mathbf{x}_D)]\right)}_{\scriptsize\boxed{3^*}}.
$$
Finally, since $\tiny\boxed{3^*}$ is independent of terms $q \in Q$ such that $w_{1,q,k} = 0$, we can reduce the size of $\tiny\boxed{1}$ by grouping the terms leading to the same value of $\tiny\boxed{3}$ together and adapting its weighting $\tiny\boxed{2}$ accordingly. Let $D^{(k)} = \{q \in D~|~w_{1,q,k}\neq0\}$. Then, $\phi_i(B, \mathbf{x})$ becomes
\begin{equation}\label{eq:shap_2}
 \underbrace{\sum_{\substack{k\in\{1,\dots,d_1\}: \\ w_{1,i,k}\neq0}}w_{2,k}}_{\scriptsize\boxed{0^*}}\underbrace{\sum_{Q\subseteq D^{(k)}\setminus \{i\}}}_{\scriptsize\boxed{1^*}} \underbrace{C(D, D^{(k)},Q)}_{\scriptsize\boxed{2^*}}\cdot\underbrace{\left(\mathbb{E}_{X_{\overline{Q\cup \{i\}}}}[L_{1,k}(\mathbf{x}_D)]-\mathbb{E}_{X_{\overline{Q}}}[L_{1,k}(\mathbf{x}_D)]\right)}_{\scriptsize\boxed{3^*}},
\end{equation}
$$
\begin{aligned}
\mbox{where } C(D, D^{(k)},Q) &= \sum_{U\subseteq D\setminus(D^{(k)}\cup\{i\})}\tfrac{(|Q|+|U|)!(|D|-|Q|-|U|-1)!}{|D|!}\\
&= \sum_{u=0}^{|D|-|D^{(k)}|-1} \tfrac{(|D|-|D^{(k)}|-1)!}{(|D|-|D^{(k)}|-1-u)!u!}\cdot\tfrac{(|Q|+u)!(|D|-|Q|-u-1)!}{|D|!}.\\
\end{aligned}
$$
Note that we also removed from $\tiny\boxed{0}$ all of the values for $k$ such that $w_{1,i,k}\neq0$, since this would lead to $\tiny\boxed{3^*}$ being 0. 

The computation of $\phi_i(B, \mathbf{x})$ in \autoref{eq:shap_2} is exponential only in the maximum number of features (non-zero weights) per hidden neuron, denoted $d_{0}^*$, which remains tractable for sparse and parsimonious BANNs. Our method for computing those values, denoted 1-BANN SHAP (see \autoref{algo:1-bann-shap}), efficiently computes the SHAP values of every feature on a whole dataset of size $m$ in $O\big(m ^ 2 \cdot d \cdot d_1 \cdot 2 ^ {d_{0}^*}\big)$. 

\begin{algorithm}[t]
	\caption{1-BANN SHAP}\label{algo:1-bann-shap}
	\begin{algorithmic}[1]
	\STATE \textbf{Input} : $S \,{=}\, \{(\mathbf{x}_j,y_j)\}_{j=1}^m$
	\STATE $\qquad\quad$\hspace{0.5mm} $B$, a BANN, with hidden weights $\mathbf{W}_1$ and output weights $\mathbf{w}_l$
        \STATE $\mathbf{R} = \mathbf{0}_{d \times d_1}$, $D = \{1,\dots,d\}$
        \STATE \textbf{For} $i \in D$ such that $\left(\exists k~|~w_{1,i,k} \neq 0 \right)$:\hfill \boxed{0^*}
        \STATE $\quad$ \textbf{For} $k \in \{1,\dots,d_1\}$ s.t. $w_{1,i,k} \neq 0$:
        \STATE $\quad$ $\quad$ $D^{(k)} = \{q \in D~|~w_{1,q,k}\neq0\}$
	    \STATE $\quad$ $\quad$ \textbf{For} $Q \subseteq D^{(k)} \setminus\{i\}$: \hfill \boxed{1^*}
        \STATE $\quad$ $\quad$ $\quad$ $C' = C(D, D^{(k)}, Q)$\hfill \boxed{2^*}
        \STATE $\quad$ $\quad$ $\quad$ \textbf{For} $\mathbf{x},\mathbf{x}' \in S$ : \hfill \boxed{3^*}
        \STATE $\quad$ $\quad$ $\quad$ $\quad$ $r_{i,k} = r_{i,k} + \frac{|w_{2,k}|C'}{m}\cdot\left|L_{1,k}\left(\mathbf{x}_{Q\setminus\{i\}}\cup \mathbf{x}'_{\overline{Q\setminus\{i\}}}\right) - L_{1,k}(\mathbf{x}_{Q}\cup \mathbf{x}'_{\overline{Q}}))\right|$ \hfill \boxed{3^*}
        \STATE $\mathbf{Return}~\mathbf{R}$
	\end{algorithmic} 
\end{algorithm}

Note that \autoref{algo:1-bann-shap} focuses on SHAP importance (SI) rather than individual SHAP values; the former is obtained by averaging each absolute value of the latter across a dataset for a given feature, thus reflecting the global contribution of a feature to the predictions of the training datasets. Also, \autoref{algo:1-bann-shap} does not directly return SIs, but a $d \times d_1$ matrix $\mathbf{R}$. 
The element $r_{i,k}$ of this matrix contains the relative importance of the feature $i$ for the hidden neuron $k$. Therefore, $\mathbf{R}$ gives the relative importance of the connection of each hidden neuron. The SI of feature $i$ is given by $\sum_{k=1}^{d_1} r_{i,k}$.

Finally, note that one can compute the relative importance of each hidden neuron themselves by doing the following: assuming a dataset \mbox{$S' = \{(L_1(\mathbf{x}_i),\mathbf{y}_i)\}_{i=1}^m$} and computing the SIs of a linear regressor (with LinearSHAP \cite{DBLP:conf/nips/LundbergL17}, a truly efficient way to do so) with parameters $(\mathbf{w}_2, b_2)$. In the following section, we illustrate how all of this information (the feature, connection, and neuron importance) can enhance the (post-hoc) explanation of a model.

%% file: sections/experiments.tex
We conduct three sets of experiments. The first one evaluates whether BGN is competitive with other techniques from the literature for training BANNs in terms of predictive accuracy on regression tasks. The second one evaluates whether pruning BANNs trained with algorithms from the literature can achieve better performances than those of BGN for an equivalent amount of parameters. The last one aims to verify whether BGN yields predictors at least as interpretable as regression trees for a similar level of accuracy.

\subsection{Predictive accuracy experiments}
\label{subsec:perf}

The first benchmark we compare BGN to is a SGD training scheme with the help of the straight-through estimator \cite{DBLP:journals/corr/BengioLC13} (BNN$^*$). We also compare to BNN+~\cite{DBLP:journals/corr/abs-1812-11800}, which was inspired by Binarized Neural Network \cite{DBLP:conf/nips/HubaraCSEB16} but makes use of a refined version of the straight-through estimator. The next two benchmarks use different continuous binarization tricks: Quantization Networks (QN) \cite{DBLP:conf/cvpr/YangSXTLDH019} and Bi-real net$^*$~\cite{DBLP:conf/eccv/LiuWLYLC18} (adapted to fully-connected architectures). Note that test performances for continuous binarization algorithms are calculated with the use of binary activation functions. We finally present the results obtained by BinaryConnect \cite{DBLP:conf/nips/CourbariauxBD15}, as often seen in the BANN training algorithm's literature. For each benchmark, three different widths (100, 500, and 1000 neurons) and depths (1, 2, and 3 hidden layers) were tested. More details about the experiment setup and the hyperparameters can be found in the appendices (\autoref{subsec:num}).

We evaluate the BGN algorithm with both \autoref{improv:1} and \autoref{improv:2}. Note that BGN has no hyperparameter tuning whatsoever since its depth is capped at 3, its width at 1000, and that the regularization parameter of the Lasso regression it uses is set such that the maximum number of features (non-zero weights) per hidden neuron ($d_0^*$) is at most 2.

\renewcommand{\arraystretch}{1.25}

\begin{table*}[t]
    \centering
    \caption{Experiment results over 3 repetitions for selected models: MSE obtained on the test dataset (Error$_{T}$) of the selected predictor, width ($d^*$), and number of hidden layers (depth - 1, $\tilde{l}$). The lowest MSE per dataset is bolded.}
    \label{tab:results_binary}
    \setlength{\tabcolsep}{1pt}
    {\tiny
    \begin{center}
    \begin{tabular}{|l||ccc||ccc|ccc|ccc|ccc|ccc|}
    \hline
    \multirow{2}{*}{Dataset} & \multicolumn{3}{c||}{BGN} & \multicolumn{3}{c|}{BC} & \multicolumn{3}{c|}{BNN$^*$} & \multicolumn{3}{c|}{BNN+} & \multicolumn{3}{c|}{Bi-real net$^*$} & \multicolumn{3}{c|}{QN}\\
    \cline{2-19}
    & Error$_T$ & $d^*$ & $\tilde{l}$ & Error$_T$ & $d^*$ & $\tilde{l}$ & Error$_T$ & $d^*$ & $\tilde{l}$ & Error$_T$ & $d^*$ & $\tilde{l}$ & Error$_T$ & $d^*$ & $\tilde{l}$ & Error$_T$ & $d^*$ & $\tilde{l}$\\
    \hline
    \hline
    bike hour \cite{bike} & 3573.67 & 259 & 2 & 21881.79 & 1000 & 3 & 1948.73 & 1000 & 3 & \textbf{1712.45} & 1000 & 2 & 1818.36 & 1000 & 3 & 2008.47 & 1000 & 2\\
    \rowcolor{Gray}
    diabete \cite{efron2004least} & \textbf{3755.93} & 6 & 1 & 9134.43 & 500 & 1 & 8919.54 & 1000 & 1 & 8999.32 & 500 & 2 & 8816.83 & 100 & 1 & 8811.23 & 500 & 2\\
    housing \cite{HARRISON197881} & \textbf{0.32} & 158 & 1 & 1.65 & 100 & 2 & 2.19 & 500 & 1 & 2.09 & 1000 & 1 & 2.36 & 500 & 1 & 2.16 & 500 & 1\\
    \rowcolor{Gray}
    hung pox \cite{rozemberczki2021chickenpox} & 223.76 & 3 & 1 & 235.75 & 100 & 2 & 250.07 & 100 & 2 & 229.83 & 100 & 3 & 242.55 & 100 & 3 & \textbf{214.27} & 500 & 3\\
    ist. stock \cite{Akbilgic2014ANH} & 2.27 & 5 & 1 & 22787.86 & 100 & 1 & 2.98 & 1000 & 3 & 1.95 & 1000 & 1 & 2.42 & 1000 & 2 & \textbf{1.94} & 1000 & 1\\
    \rowcolor{Gray}
    parking \cite{10.1007/978-3-319-59513-9_11} & \textbf{6954.87} & 1000 & 1 & 176286.81 & 1000 & 1 & 32705.94 & 500 & 3 & 11684.12 & 1000 & 3 & 16543.25 & 1000 & 2 & 107347.98 & 1000 & 2\\
    power p.\cite{TUFEKCI2014126,HeysemKayaLocalAG} & 17.18 & 187 & 2 & 13527.9 & 500 & 2 & 18.34 & 1000 & 1 & \textbf{16.66} & 1000 & 2 & 16.77 & 1000 & 2 & 17.14 & 1000 & 2\\
    \hline
    \end{tabular}
    \end{center}}
\end{table*}

We experiment with a variety of datasets having different numbers of examples (315 to 36~736) and features (4 to 20). The architecture and hyperparameter set yielding the best average validation MSE over 3 random initialization and train/validation separation are selected and reported in \autoref{tab:results_binary}, as well as their mean test MSE.
The results clearly show that BGN is competitive on various regression tasks when compared to several algorithms from the literature for training BANNs. On some datasets, all of the baseline methods surprisingly fail to provide a good predictor, despite all of the hyperparameter combinations that were tested (between 27 and 324 combinations per dataset per benchmark). The most surprising result occurs on the \textit{diabete} dataset, where BGN's test MSE is more than two times smaller than the best runner-up. We hypothesize that existing BANN algorithms were not tested in the regression on tabular data regimes.

\autoref{tab:results_binary} also highlights that the predictors given by BGN are consistently shallower and narrower than those selected by the baselines; even though its BANNs could grow layers of 1000 neurons, BGN usually converges way before it happens. On simple tasks, BGN achieves competitive prediction using very few parameters, whereas the baselines, to obtain the best performances possible, must contain up to millions of parameters. Therefore, even when interpretability is not considered, BGN competes with state-of-the-art approaches for empirical performances for regression on tabular data while yielding simple predictors.

\subsection{Pruning experiments}\label{subsec:pruning}

\begin{figure}[t]
\begin{minipage}[c]{0.48\textwidth}
\begin{center}
\includegraphics[width=1\linewidth]{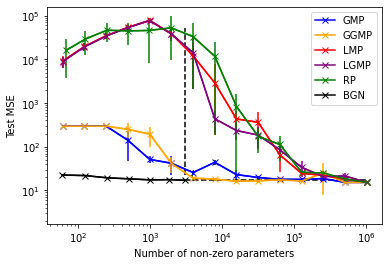} \caption{Pruning of BNN+ network on the \textit{Power Plant} dataset over 5 random seeds. The dotted lines show where BGN has converged. Vertical lines depicts one standard deviation.} \label{fig:pruning1}
\end{center}
\end{minipage}
\begin{minipage}[c]{0.04\textwidth}
\quad
\end{minipage}
\begin{minipage}[c]{0.48\textwidth}
\begin{center}
\includegraphics[width=1\linewidth]{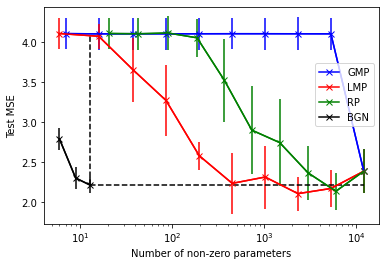} \caption{Pruning of QN network on the \textit{Istanbul Stock USD} dataset. The dotted lines show where BGN has converged. Vertical lines depicts one standard deviation.} \label{fig:pruning2}
\end{center}
\end{minipage}
\end{figure}

The first set of experiments showed that BGN can yield truly small yet accurate predictors. We now check whether it is possible to prune BANNs obtained by the benchmarks to obtain simpler yet better-performing networks than those of BGN. We experiment on two datasets from \autoref{subsec:perf} for which BGN is outperformed by a benchmark: \textit{Istanbul Stock USD} and \textit{Power Plant}. We then select the benchmark algorithm that resulted in the best performances over three different random seeds in both of these tasks (respectively QN and BNN+), with their corresponding set of hyperparameters. We iteratively pruned half the non-zero weights of the trained network 
thanks to five state-of-the-art pruning methods  \cite{DBLP:journals/corr/abs-1903-01611}\cite{DBLP:journals/corr/abs-1902-09574}\cite{DBLP:conf/nips/HanPTD15}\cite{DBLP:journals/corr/HanMD15}: Global Magnitude Pruning (GMP), Layerwise Magnitude Pruning (LMP), Global Gradient Magnitude Pruning (GGMP), Layerwise Gradient Magnitude Pruning (LGMP).  We also included Random Pruning (RP).
At each iteration, after the pruning phase, the network's remaining weights are fine-tuned by using the same learning algorithm as in the original training phase (QN or BNN+), but the learning rate is divided by 10 at each iteration.
We report the MSE on the test set (on 5 different random seeds) over the pruning operations and compared it to the test MSE obtained by BGN while building its predictor. We report those quantities as a function of the number of non-zero weights, a common metric for comparing pruning methods \cite{DBLP:conf/mlsys/BlalockOFG20}.

It seems that obtaining, from the baselines, a compact but efficient predictor is not as easy as training a huge BANN and then pruning it. \autoref{fig:pruning1} and \autoref{fig:pruning2} both are in line with that assertion: pruning the best predictor obtained from the best baseline until it contains as few non-zero parameters as BGN's BANNs truly jeopardizes their predictive performances. Baselines and BGN are comparable solely in terms of predictive performances; when it comes to efficiency, BGN shines, certainly because this algorithm was conceived to obtain compact predictors from the beginning.

\subsection{Interpretability and explainability of 1-BANNs}\label{subsec:intexp}

\begin{figure}[t]
\begin{center}
\includegraphics[width=0.9\linewidth]{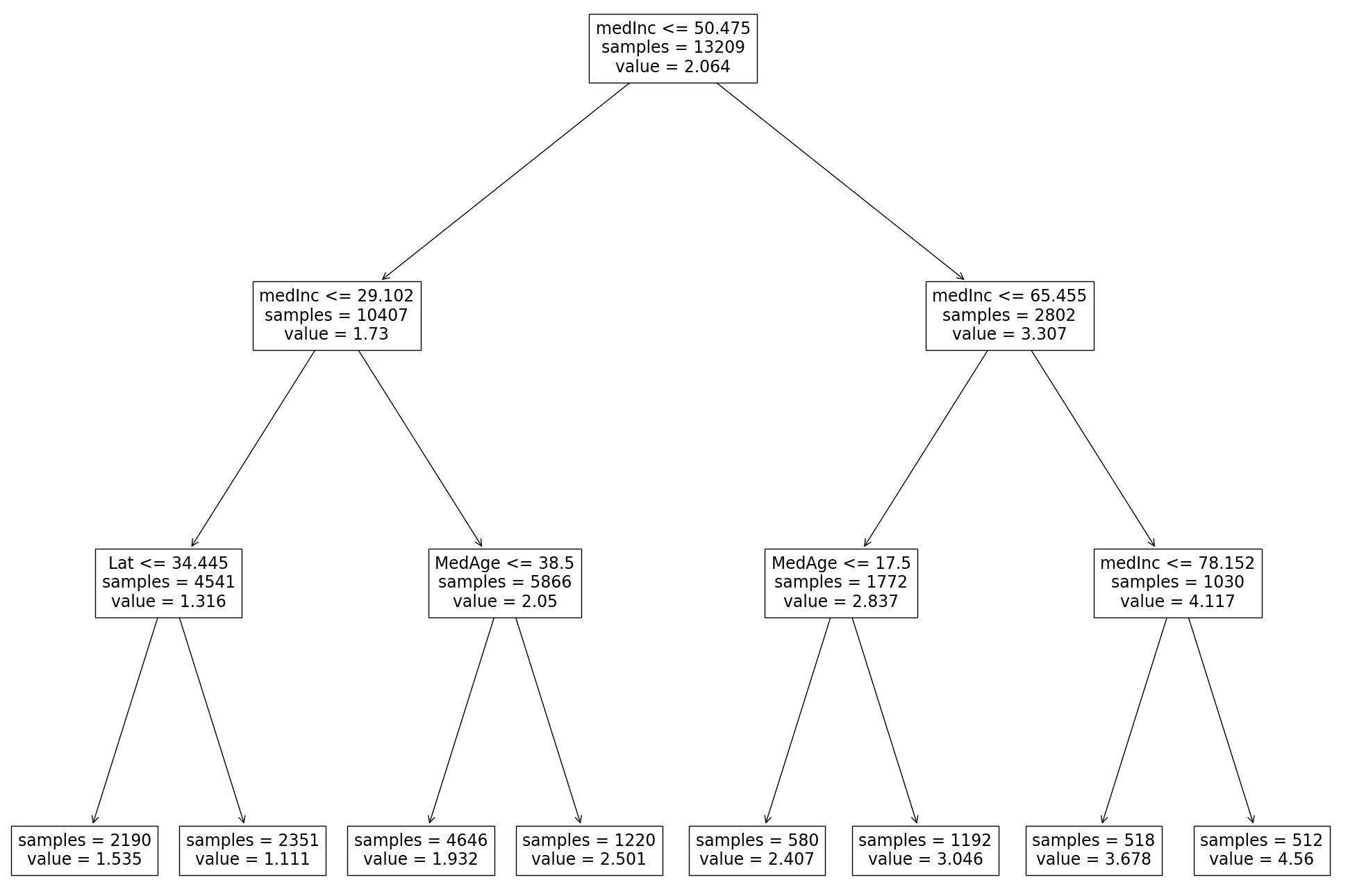} \caption{A visual representation of a regression tree of depth three trained on the \textit{housing} dataset, created by the scikit-learn library \cite{pedregosa2011scikit}.} \label{fig:tree}
\end{center}
\end{figure}

In order to study the interpretability assets of a BANN learned by BGN, we compare the obtained predictor to the one of a decision tree for the same level of accuracy on the \textit{housing} dataset. The latter is a regression problem where the goal is to predict the cost of a house, in hundreds of thousands of US dollars. The dataset contains eight features. More precisely, we train a regression tree of depth three on the \textit{housing} dataset (depicted on \autoref{fig:tree}); then, we use BGN for growing a 1-BANN until it reaches at least the accuracy of the regression tree. Knowing that a depth-3 regression tree is quite interpretable, we now analyze the obtained 1-BANN predictor to see if this one also has interesting interpretability and explainability properties for equal predictive performances. More details on the regression tree training can be found in appendices (\autoref{sec:int-expl}).

Out of the eight input features, only three were retained by the BGN predictor: median age of a house within a block (MedAge); total number of bedrooms within a block (TotalBed); median income for households within a block of houses, measured in thousands of US Dollars (MedInc). We present the obtained BANN predictor in two different ways: the first one emphasizing its interpretable aspect (\autoref{fig:pred-int}, its mathematical equation), and the second one its explainable aspect (\autoref{fig:pred-expl}, a visual depiction). The predictor is presented with threshold activation, for interpretability's sake. The predictor has only one hidden layer of width $5$.

\definecolor{col-1}{RGB}{0,0,255}
\definecolor{col-2}{RGB}{102, 51, 0}
\definecolor{col-3}{RGB}{38,185,8}
\definecolor{col-4}{RGB}{204,0,0}
\definecolor{col-5}{RGB}{215, 215, 0}

\begin{figure}[t]
\begin{tcolorbox}
\small
\vspace{-.4cm}
\begin{align*}
B(_{\textup{MedInc, MedAge, TotalBed}}) = 1.00~&+ \textbf{\textcolor{col-4}{1.27}} \cdot \mathbbm{1}_{\{\textup{MedInc} > 65.46\}} + \\
\hspace{-.2cm}\textbf{\textcolor{col-5}{1.01}} \cdot \mathbbm{1}_{\{0.59 \cdot \textup{MedAge} + \textup{MedInc} > 63.56\}} &+ \textbf{\textcolor{col-3}{0.60}} \cdot \mathbbm{1}_{\{\textup{MedInc} > 28.20\}} + \\\textbf{\textcolor{col-1}{0.36}} \cdot \mathbbm{1}_{\{\textup{TotalBed} > 622.0\}} &+ \textbf{\textcolor{col-2}{0.27}} \cdot \mathbbm{1}_{\{\textup{MedAge} > 20.0\}}
\end{align*}
\end{tcolorbox}
\caption{The predictive model of 1-BANN predictor trained on the \textit{housing} dataset.}
\label{fig:pred-int}
\end{figure}

\tikzstyle{neuralnetwork}=[%
draw=black,
node distance = 6mm and 25mm,
start chain = going right,
neuron/.style = {rectangle, draw=black,
	minimum size=16pt, inner sep=4pt,
	on chain},
annot/.style = {text width=4em, align=center,},
tips]

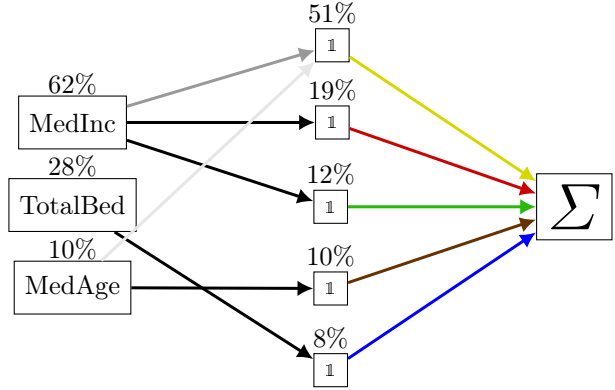
\begin{figure}[t]
\begin{center}
\begin{tikzpicture}[neuralnetwork]
	\def\layersd{7.4mm}
	\def\neuronsize{20pt}
	\def\edgethickness{0.4mm}
	
        \node[neuron, minimum size=0.1\neuronsize] (H1-2) {\scriptsize $\mathbbm{1}$};

        \node[above=1.8mm of H1-2.center] (H1.5-2) {\normalsize 8\%};
    
	\node[neuron,above left=\layersd and 26.5mm of H1-2.center, minimum size=\neuronsize]  (H2-1)   {\normalsize MedAge};
        \node[neuron, above=8.6mm of H1-2.center, minimum size=0.1\neuronsize] (H2-2)  {\scriptsize $\mathbbm{1}$};

        \node[above=2.6mm of H2-1.center] (H2.5-1) {\normalsize 10\%};
        \node[above=1.8mm of H2-2.center] (H2.5-2) {\normalsize 10\%};

        \node[neuron,above=\layersd of H2-1.center, minimum size=\neuronsize]  (H3-1)   {\normalsize TotalBed};
        \node[neuron, above=8.6mm of H2-2.center, minimum size=0.1\neuronsize] (H3-2)  {\scriptsize $\mathbbm{1}$};
        \node[neuron, minimum size=\neuronsize] (H3-3)  {$\Sigma$};

        \node[above=2.6mm of H3-1.center] (H3.5-1) {\normalsize 28\%};
        \node[above=1.8mm of H3-2.center] (H3.5-2) {\normalsize 12\%};
        
        \node[neuron,above=\layersd of H3-1.center, minimum size=\neuronsize]  (H4-1)   {\normalsize MedInc};
        \node[neuron, minimum size=0.1\neuronsize] (H4-2)  {\scriptsize $\mathbbm{1}$};

        \node[above=2.6mm of H4-1.center] (H4.5-1) {\normalsize 62\%};
        \node[above=1.8mm of H4-2.center] (H4.5-2) {\normalsize 19\%};

        \node[neuron,above=8mm of H4-2.center, minimum size=0.1\neuronsize]  (H5-2)   {\scriptsize $\mathbbm{1}$};

        \node[above=1.8mm of H5-2.center] (H5.5-2) {\normalsize 51\%};


    \path[line width=\edgethickness, color=col-1](H1-2) edge (H3-3);
    \path[line width=\edgethickness, color=col-2](H2-2) edge (H3-3);
    \path[line width=\edgethickness, color=col-3](H3-2) edge (H3-3);
    \path[line width=\edgethickness, color=col-4](H4-2) edge (H3-3);
    \path[line width=\edgethickness, color=col-5](H5-2) edge (H3-3);
    \definecolor{Gray}{gray}{0.}
    \path[line width=\edgethickness, color=Gray](H4-1) edge (H4-2);

    \definecolor{Gray}{gray}{0.}
    \path[line width=\edgethickness, color=Gray](H4-1) edge (H3-2);
    
    \definecolor{Gray}{gray}{0.}
    \path[line width=\edgethickness, color=Gray](H2-1) edge (H2-2);

    \definecolor{Gray}{gray}{0.}
    \path[line width=\edgethickness, color=Gray](H3-1) edge (H1-2);

    \definecolor{Gray}{gray}{0.9}
    \path[line width=\edgethickness, color=Gray](H2-1) edge (H5-2);%
    \definecolor{Gray}{gray}{0.6}
    \path[line width=\edgethickness, color=Gray](H4-1) edge (H5-2);%
\end{tikzpicture}

\caption{Visual representation of a 1-BANN for the \textit{housing} dataset. The values over the input neurons and hidden neurons correspond to their respective SHAP values, at a predictor level. The blackness of the weights connecting the input layer to the hidden layer is proportional to the SHAP value, at a neuron level.}
\label{fig:pred-expl}
\end{center}
\end{figure}

\autoref{fig:pred-int} presents the equation describing the 1-BANN predictor. To simplify the equation, all of the colored coefficients (output weights) are listed in decreasing order and when possible the coefficients associated with input features are set to one by scaling the threshold accordingly. Since the hidden neurons (thresholds, in the equation) are independent, their role in the prediction is easy to understand, as well as what could have been different in the input of a given neuron to change its output value. 
Both the base prediction value (here, 1.00) and possible interventions on the features to increase the predicted outcome are explicit, thanks to the additive form of the predictor (a property that does not stand for regression trees). Thus, simple BANNs as those outputted by BGN truly are interpretable thanks to their modular decomposition and lightness. 

\autoref{fig:pred-expl} presents a visual depiction of the network predictor. Three types of relative SHAP importance ($rSI$, where $rSI_i = SI_i / (\sum_j SI_j)$) are present: two at the predictor level (feature values, above the features; hidden neuron values, above the hidden neurons), and one at a neuron level (the color intensity of a connection is proportional to its neuron-wise $rSI$). We display the features and hidden neurons in descending order according to their $rSI$, making the top features and neurons the most relevant to the predictions. For example, \textit{MedInc} is especially important. There is only one neuron connected to two input features, whose output is more impacted by the \textit{MedInc} feature than \textit{TotalBed} one. 

Also, because of the color code linking both the representations from \autoref{fig:pred-expl} and \autoref{fig:pred-int}, it is easy to bridge the information of both representations together. In the appendices (\autoref{subsec:intexp}), we present how, as the complexity of the regression tree grows exponentially (in its depth), BGN only needs to grow the hidden layer a bit in order to keep up with the gain in predictive performances of the tree.

%% file: sections/conclusion.tex
We provided a comprehensive study of binary activated neural networks (BANNs) and leveraged this knowledge to create an algorithm that has solid theoretical groundings for efficiently training highly interpretable BANNs. To better explain the relative importance of the various components of these predictors, we extended the original algorithm for computing SHAP values for features to hidden neurons and connections. For parsimonious and sparse BANNs, the computation of the SHAP values becomes tractable. We showed that the method can efficiently tackle many regression problems, obtaining MSE comparable to state-of-the-art methods for training BANNs. We presented an analysis of a BANN trained on the \textit{housing} dataset, arguing the predictor had interpretability and explainability advantages over a similarly performing regression tree. 
Future work includes building in a similar way binary activated neural networks for multi-label classification tasks and more complicated NN architectures, like convolutional neural networks.

%% file: sections/supplementary.tex
\subsection{The Complete BGN Algorithm}\label{sec:bgn_algos}

See \autoref{algo:bgn_general} for a complete view of the algorithm, with both \autoref{improv:1} and \autoref{improv:2}.

\begin{algorithm}[h]
	\caption{The complete BGN algorithm}
	\begin{algorithmic}[1]\label{algo:bgn_general}
	\STATE \textbf{Input} : $S = \{(\xbf_1,y_1),\dots,(\xbf_m,y_m)\}$, $\ybf \in \mathbb{R}$, $\xbf\in\mathbb{R}^d$, a dataset
	\STATE Set $\{\mathbf{x}_i^{(1)}\}_{i=1}^m = \{\mathbf{x}_i\}_{i=1}^m$
	\STATE \textbf{For} $z = 1,\dots,l-1$ :
    \STATE $\quad$ Set $\mathbf{r}^{(1)} = \ybf$, $t=0$, $b_{z+1} = 0$, $\Wbf_{z} = \textbf{0}_{d_{z-1}\times d_{z}}$, $\bbf_{z} = \textbf{0}_{d_{z}}$
	\STATE $\quad$ \textbf{While} stopping criterion is not met:
	\STATE $\quad\quad$ $t = t + 1$
    \STATE $\qquad$ $\wbf_{z,t} = \textup{LassoRegression}(S^{(z,t)})$,
	\quad where $S^{(z,t)}= \{(\xbf_1^{(z)},r_1^{(t)}),\dots,(\xbf_m^{(z)},r_m^{(t)})\}$
    \STATE $\qquad$ $b_{z,t} = {\argmin}_b\ \left( \frac{|\mathbf{r}^{(z,t)}_{-1}|}{m}\textup{Var}\left(\mathbf{r}^{(z,t)}_{-1}\right)+\frac{|\mathbf{r}^{(z,t)}_{+1}|}{m}\textup{Var}\left(\mathbf{r}^{(z,t)}_{+1}\right)\right)$ \\
    \STATE $\qquad\qquad\qquad\quad$ with $\mathbf{r}^{(z,t)}_{\pm1}$ given by Eq.~\eqref{eq:ypm}, but with $\{\mathbf{x}_i^{(t)}\}_{i=1}^m$ instead of $\{\mathbf{x}_i\}_{i=1}^m$
	\STATE  $\qquad$ $\rho_\pm^{(z,t)} = \dfrac{\sum_{i:\sgn\left(\xbf_i^{(z)}\cdot\mathbf{w}_{z,t}+b_{z,t}\right)=\pm1} r_i^{(t)}}{\sum_{i:\sgn\left(\xbf_i^{(z)}\cdot\mathbf{w}_{z,t}+b_{z,t}\right)=\pm1} 1}$
	\STATE $\qquad$ $w_{z+1,t} = \frac{1}{2}\left(\rho_+^{(z,t)}-\rho_-^{(z,t)}\right); \quad  b_{z+1,t} = \frac{1}{2}\left(\rho_+^{(z,t)}+\rho_-^{(z,t)}\right)$
    \STATE $\quad\quad$ $r^{(t+1)}_{i}=r^{(t)}_{i} - w_{z+1,t}\ \sgn(\mathbf{w}_{z,t}\cdot\mathbf{x}_i^{(z)}+b_{z,t}) - b_{z+1,t}$\ \ $\forall i \in \{1,\dots,m\}$
	\STATE $\quad\quad$ $p^*=1$
	\STATE $\quad\quad$ \textbf{While} $p^* < t$:
	\STATE $\quad\quad\quad$ $p$ = random.int($1,t$)
    \STATE $\qquad\quad$ $\delta$ = $\overline{\mathbf{r}^{(p)}}$
	\STATE $\qquad\quad$ $\gamma$ = ($\wbf_{z,p}$, $b_{z,p}$, $w_{z+1,p}$, $b_{z+1,p}$, $\mathbf{r}^{(p)}$) \hfill [Save values]
    \STATE $\quad\quad\quad$ $r^{(p)}_{i}=r^{(p)}_{i} + w_{z+1,p}\ \sgn(\mathbf{w}_{z,p}\cdot\mathbf{x}_i^{(z)}+b_{z,p}) + b_{z+1,p}$\ \ $\forall i \in \{1,\dots,m\}$
    \STATE $\qquad\quad$ Do steps 8-12, with $t:=p$
	\STATE $\quad\quad\quad$ $r^{(p)}_{i}=r^{(p)}_{i} - w_{z+1,p}\ \sgn(\mathbf{w}_{z,p}\cdot\mathbf{x}_i^{(z)}+b_{z,p}) - b_{z+1,p}$\ \ $\forall i \in \{1,\dots,m\}$
    \STATE $\qquad\quad$ \textbf{If} $\overline{\mathbf{r}^{(p)}} > \delta$
    \STATE $\qquad\qquad$ ($\wbf_{z,p}$, $b_{z,p}$, $w_{z+1,p}$, $b_{z+1,p}$, $\mathbf{r}^{(p)}$) = $\gamma$ \hfill [Restore values]
    \STATE $\qquad\quad$ $p^* = p^* + 1$
	\STATE $\quad$ $\mathbf{x}_i^{(z+1)} = (L_{z}\circ\dots\circ L_1)(\mathbf{x}_i) \forall i \in \{1,\dots,m\}$
	\STATE \textbf{Output} : $BGN_T(\xbf)$ = $\sum_{t=1}^T \mathbf{c}_t\ (L_{l-1}\circ\dots\circ L_1)(\mathbf{x})+\mathbf{d}_t$
	\end{algorithmic} 
\end{algorithm}

\subsection{Details about the numerical experiments}\label{subsec:num}
\renewcommand{\arraystretch}{1.1}
\begin{table}[t]
\scriptsize
    \centering
    \caption{Datasets overview (F = Floats, I = Integers)}
    \label{tab:algo_overview}
    \setlength{\tabcolsep}{4pt}
    {\small
    \begin{tabular}{llllllll}
    \toprule
    Dataset & Full name & Taken from & Source & $d$ & $\mathbf{x}$ & $m$\\
    \midrule
 bike hour & Bike sharing dataset & UCI Rep. & \cite{bike} & 16 & F/I & 17 389 \\
     diabete & Diabetes & SKLearn & \cite{efron2004least} & 10 & F/I & 442 \\
     housing & California housing & SKLearn & \cite{HARRISON197881} & 8 & F & 20 640 \\
    hung pox & Hungarian chickenpox cases & UCI Rep. & \cite{rozemberczki2021chickenpox} & 20 & F & 521 \\
    ist. stock & Istanbul stock exchange (USD) & UCI Rep. & \cite{Akbilgic2014ANH} & 8 & F & 536 \\
    parking & Parking Birmingham & UCI Rep. & \cite{10.1007/978-3-319-59513-9_11} & 4 & F & 35 717 \\
    power p. & Combined cycle power plant & UCI Rep. & \cite{TUFEKCI2014126,HeysemKayaLocalAG} & 4 & F & 9568 \\
    \bottomrule
    \end{tabular}
    }
\end{table}

\begin{table}[t]
    \centering
    \caption{Benchmarks overview}
    \label{tab:methods_overview}
    \setlength{\tabcolsep}{3pt}
    {\small
    \begin{tabular}{l||c|ccccc}
    \cline{1-7}
    Algorithm & BGN & BC & BNN$^*$ & BNN+ & Bi-real net$^*$ & QN\\
    \cline{1-7}
    Weights & $\mathbb{R}$ & $\{-1,+1\}$ & $\mathbb{R}$ & $\mathbb{R}$ & $\mathbb{R}$ & $\mathbb{R}$\\
    Activations output & $\{-1,+1\}$ & $\mathbb{R}$ & $\{-1,+1\}$ & $\{0,1\}$ & $\{-1,+1\}$ & $\{0,1\}$\\
    Uses batch norm & False & True & True & True & True & True\\
    Uses regularization & False & False & False & True & False & False\\
    \cline{1-7}
    \end{tabular}
    }
\end{table}

\paragraph{Dataset splits}
\begin{itemize}
    \item Test set proportion: 25\% of the total dataset; validation set: 20\% of the remaining data.
\end{itemize}
    
\paragraph{Hyperparameters selected on the validation set}
\begin{itemize}
    \item Number of hidden layers: 1,2,3
    \item Width: 100, 500, 1000
    \item Learning rates: 0.1, 0.01, 0.001
    \item Regularization type (BNN+): $L_1$, $L_2$ ($10^{-6}$, $10^{-7}$)
    \item $\beta$ values (BNN+): 1, 2, 5
    \item $T_{\textup{Start}}$ values (QN): 5, 10, 20 (while $T_{\textup{At epoch n}}$ = $T_{\textup{Start}} * n$)
\end{itemize}

\paragraph{Fixed hyperparameters}
\vspace{-2mm}
\begin{itemize}
    \item Initialization: Kaiming uniform  \cite{DBLP:conf/iccv/HeZRS15}
    \item Batch size: 512 for big datasets (over 9000 examples) and 64 for small ones (see \autoref{tab:algo_overview})
    \item Maximum number of epochs: 200 (early stop: 20)
    \item Optimization algorithm: Adam \cite{DBLP:journals/corr/KingmaB14},\\ with $\epsilon = 0.001, \rho_1 = 0.9, \rho_2 = 0.999, \delta = 10^{-8}, 
    \lambda=0 $
    \item Learning rate decay: plateau (patience: 5)
\end{itemize}

\paragraph{Computation time}

\begin{table}[b]
    \centering
    \caption{Training time of BGN on some experimental datasets (max. depth: 3; max. width: 1000).}
    \label{tab:time}
    \begin{tabular}{|l|c|}
    \hline
        Dataset & Training time (minutes)\\
    \hline
        Hungarian chickenpox & 0.03 $\pm$ 0.01 \\
        California housing & 156.47 $\pm$ 25.89 \\
        Bike sharing dataset & 628.31 $\pm$ 45.19 \\
    \hline
    \end{tabular}
\end{table}

\begin{itemize}
    \item See \autoref{tab:time} for examples of training time of BGN on various datasets used in \autoref{section:experiments}.
\end{itemize}

\subsection{Extended details on the interpretability and explainability experiments}\label{sec:int-expl}

The experiments were conducted on five random seeds, for BGN as well as for the regression tree. As for the tree, the following hyperparameter choices were considered (with the retained set of hyperparameters being bolded): maximum number of features (\textit{d}) considered at each split: \textbf{d}, $\sqrt{d}$, $log_2(d)$; criterion: \textbf{squared error}, Friedman MSE, absolute error, poisson; the strategies used to choose the split at each node: random, \textbf{best}.

\begin{table}[t]
\caption{Comparison between trees of depth 3, 4, and 5 and 1-BANNs having correspondingly similar mean test MSE on the \textit{housing} dataset. We present the tree depth and the 1-BANN width ($d_1$) as a metric for comparing their complexity and the number of considered features of the models ($d_0^*$).}
\centering
\setlength{\tabcolsep}{2pt}
\begin{tabular}{|c c|cc|} 
 \hline
 Model & Test MSE & (Depth, $d_1$) & $d_0^*$ \\
 \hline\hline
 1-BANN & 0.6358 & 5 & 3 \\
 Tree & 0.6413 & 3 & 3 \\ 
 \hline
 1-BANN & 0.5849 & 7 & 4 \\
 Tree & 0.5779 & 4 & 5 \\
 \hline
 1-BANN & 0.5020 & 8 & 5 \\
 Tree & 0.5291 & 5 & 6 \\
 \hline
\end{tabular}
\label{tab:perf-compl}
\end{table}

\autoref{tab:perf-compl} shows how, on the \textit{housing} dataset, as trees become deeper, thus exponentially more powerful and complex, BGN only has to yield 1-BANNs having a few more hidden neurons to keep up with the gains in predictive performances. We see that the required width of the 1-BANNs grows almost linearly, and while the depth of the tree has a huge impact on its interpretability, the width of the 1-BANN has a relationship to its interpretability that is way less important (as analyzed in \autoref{subsec:intexp}).

%% file: sections/math.tex
\label{sec:proofs}


\begin{proof}[\autoref{prop:bin_bnd}]
\vspace{-2mm}
$$
\begin{aligned} \allowdisplaybreaks
\ell_S(B) &= \frac{1}{m}\sum_{(\xbf, y)\in S} \left(B(\xbf) - y\right)^2 = \frac{1}{m}\sum_{(\xbf, \mathbf{y})\in S} (\mathbf{w}_{l}\cdot L_{1:l-1}(\mathbf{x}) + b_{l} - y)^2\\
&= \frac{1}{m}\sum_{\pbf\in\mathcal{L}_{1:l-1}}\sum_{\substack{(\xbf, \mathbf{y})\in S: \\ L_{1:l-1}(\mathbf{x})=\mathbf{p}}}(\mathbf{w}_{l}\cdot\pbf + b_{l} - y)^2\\
&\geq \frac{1}{m}\sum_{\pbf\in\mathcal{L}_{1:l-1}}\argmin_{\mathbf{w},b}\sum_{\substack{(\xbf, \mathbf{y})\in S: \\ L_{1:l-1}(\mathbf{x})=\mathbf{p}}}(\mathbf{w}\cdot\pbf + b - y)^2\\
&= \sum_{\pbf\in\mathcal{L}_{1:l-1}}\frac{1}{m}\sum_{\substack{(\xbf, \mathbf{y})\in S \\ L_{1:l-1}(\mathbf{x})=\mathbf{p}}}\left(\overline{\ybf_{\pbf}^{(l-1)}} - y\right)^2 = \sum_{\pbf\in\mathcal{L}_{1:l-1}}\frac{\left|\ybf_{\pbf}^{(l-1)}\right|}{m}\textup{Var}\left(\ybf_{\pbf}^{(l-1)}\right)\ ,
\end{aligned}
$$
where $\overline{\ybf} = \frac{1}{m}\sum_{y\in\ybf}y$. We saw in \autoref{section:BANN} that hidden layers $L_2$ to $L_l$ group regions created by the preceding hidden layer. Using the fact that for two datasets $S_1$ and $S_2$,
$$
\begin{aligned}
&\min_{\mathbf{w},b} \left(\sum_{(\xbf, y)\in S_1} (\mathbf{w}\cdot\mathbf{x}+b-y_i)^2 + \sum_{(\xbf, y)\in S_2} (\mathbf{w}\cdot\mathbf{x}+b-y_i)^2\right) \\
\geq &\min_{\mathbf{w}_1,b_1} \sum_{(\xbf, y)\in S_1} (\mathbf{w}_1\cdot\mathbf{x}+b_2-y_i)^2 + \min_{\mathbf{w}_2,b_2} \sum_{(\xbf, y)\in S_2} (\mathbf{w}_2\cdot\mathbf{x}+b_2-y_i)^2\ ,
\end{aligned}
$$
we obtain that $\forall k \in \{2,\dots,l\}$: 
$$
\sum_{\pbf\in\mathcal{L}_{1:k}}\sum_{\substack{(\xbf, \mathbf{y})\in S \\ L_{1:k}(\mathbf{x})=\mathbf{p}}}\left(\overline{\ybf_{\pbf}^{(k)}} - y\right)^2\geq\sum_{\pbf\in\mathcal{L}_{1:k-1}}\sum_{\substack{(\xbf, \mathbf{y})\in S \\ L_{1:k-1}(\mathbf{x})=\mathbf{p}}}\left(\overline{\ybf_{\pbf}^{(k-1)}} - y\right)^2\ ,
$$
completing the proof.\qed
\end{proof}


\begin{proof}[{\autoref{prop:BGN_min}}] Note that $\frac{\partial}{\partial w}\mathbf{w}_{1,t} = \frac{\partial}{\partial b}b_{1,t} = 0$
$$
\begin{aligned}
\allowdisplaybreaks
\frac{\partial}{\partial w}&\sum_i \big(r_i^{(t)}-w\ \sgn(\mathbf{w}_{1,t}\cdot\xbf_i+b_{1,t})-b\big)^2\\
&= -2\sum_i \left(r_i^{(t)}-w\ \sgn(\mathbf{w}_{1,t}\cdot\xbf_i+b_{1,t})-b\right)\sgn(\mathbf{w}_{1,t}\cdot\xbf_i+b_{1,t})\\
\Rightarrow 0 &= -2\sum_i \left(r_i^{(t)}-\hat{w}\ \sgn(\mathbf{w}_{1,t}\cdot\xbf_i+b_{1,t})-\hat{b}\right)\sgn(\mathbf{w}_{1,t}\cdot\xbf_i+b_{1,t})\\
\Rightarrow 0 &= \sum_i r_i^{(t)}\sgn(\mathbf{w}_{1,t}\cdot\xbf_i+b_{1,t})-n\hat{w}-\sum_i \hat{b}\cdot\sgn(\mathbf{w}_{1,t}\cdot\xbf_i+b_{1,t})\\
\Rightarrow \hat{w} &= \frac{1}{n}\sum_i (r_i^{(t)}-\hat{b})\cdot\sgn(\mathbf{w}_{1,t}\cdot\xbf_i+b_{1,t})\\
\end{aligned}
$$

$$
\begin{aligned}
\frac{\partial}{\partial b}\sum_i \left(r_i^{(t)}-w\ \sgn(\mathbf{w}_{1,t}\cdot\xbf_i+b_{1,t})-b\right)^2 &= -2\sum_i \left(r_i^{(t)}-w\ \sgn(\mathbf{w}_{1,t}\cdot\xbf_i+b_{1,t})-b\right)\\
\Rightarrow 0 &= -2\sum_i \left(r_i^{(t)}-\hat{w}\ \sgn(\mathbf{w}_{1,t}\cdot\xbf_i+b_{1,t})-\hat{b}\right)\\
\Rightarrow \hat{w} &= \frac{\sum_i r_i^{(t)}-n\hat{b}}{\sum_i \sgn(\mathbf{w}_{1,t}\cdot\xbf_i+b_{1,t})}\\
\end{aligned}
$$
In order to lighten the notation, we write $\sum_{\pm1}$ to denote $\sum_{i:\sgn(\mathbf{w}_{1,t}\cdot\xbf_i+b_{1,t})=\pm1}$.
$$
\begin{aligned}
\frac{1}{n}\sum_i (r_i^{(t)}-\hat{b})\sgn(\mathbf{w}_{1,t}\cdot\xbf_i+b_{1,t}) &= \frac{\sum_i r_i^{(t)}-n\hat{b}}{\sum_i \sgn(\mathbf{w}_{1,t}\cdot\xbf_i+b_{1,t})}\\
\Rightarrow \frac{(\sum_{+1}r_i^{(t)}-\sum_{-1}r_i^{(t)})-\hat{b}(\sum_{+1}1-\sum_{-1}1)}{\sum_{+1}1+\sum_{-1}1} &= \frac{(\sum_{+1}r_i^{(t)}+\sum_{-1}r_i^{(t)})-\hat{b}(\sum_{+1}1+\sum_{-1}1)}{\sum_{+1}1-\sum_{-1}1}\\
\Rightarrow \hat{b}[(\sum_{+1}1+\sum_{-1}1)^2-(\sum_{+1}1-\sum_{-1}1)^2] &= (\sum_{+1}r_i^{(t)}+\sum_{-1}r_i^{(t)})(\sum_{+1}1+\sum_{-1}1)\\
&-(\sum_{+1}r_i^{(t)}-\sum_{-1}r_i^{(t)})(\sum_{+1}1-\sum_{-1}1)\\
\Rightarrow 4\hat{b}\sum_{+1}1\sum_{-1}1 &= 2\sum_{+1}r_i^{(t)}\sum_{-1}1+2\sum_{-1}r_i^{(t)}\sum_{+1}1\\
\Rightarrow \hat{b} &= \frac{1}{2}\left(\frac{\sum_{+1}r_i^{(t)}}{\sum_{+1}1}+\frac{\sum_{-1}r_i^{(t)}}{\sum_{-1}1}\right)
\end{aligned}
$$

\begin{eqnarray*}
\hat{w} &=& \frac{\sum_i r_i^{(t)}-n\hat{b}}{\sum_i \sgn(\mathbf{w}_{1,t}\cdot\xbf_i+b_{1,t})}\\
\Rightarrow \hat{w} &=& \frac{\sum_{+1} r_i^{(t)}+\sum_{-1} r_i^{(t)}}{\sum_{+1}1-\sum_{-1}1}-\hat{b}\frac{\sum_{+1}1+\sum_{-1}1}{\sum_{+1}1-\sum_{-1}1}\\
\Rightarrow \hat{w} &=& \frac{\sum_{+1} r_i^{(t)}+\sum_{-1} r_i^{(t)}}{\sum_{+1}1-\sum_{-1}1}-\frac{1}{2}\left(\frac{\sum_{+1}r_i^{(t)}}{\sum_{+1}1}+\frac{\sum_{-1}r_i^{(t)}}{\sum_{-1}1}\right)\frac{\sum_{+1}1+\sum_{-1}1}{\sum_{+1}1-\sum_{-1}1}\\
\Rightarrow \hat{w}(\sum_{+1}1-\sum_{-1}1) &=& \sum_{+1}r_i^{(t)}+\sum_{-1}r_i^{(t)}-\frac{1}{2}\frac{\sum_{+1}r_i^{(t)}\sum_{-1}1+\sum_{-1}r_i^{(t)}\sum_{+1}1}{\sum_{+1}1\sum_{-1}1}(\sum_{+1}1+\sum_{-1}1)\\
\Rightarrow \hat{w}(\sum_{+1}1-\sum_{-1}1) &=& \frac{1}{2}\sum_{+1}r_i^{(t)}+\frac{1}{2}\sum_{-1}r_i^{(t)}-\frac{1}{2}\frac{\sum_{+1}r_i^{(t)}\sum_{-1}1}{\sum_{+1}1}-\frac{1}{2}\frac{\sum_{-1}r_i^{(t)}\sum_{+1}1}{\sum_{-1}1}\\
\Rightarrow \hat{w}(\sum_{+1}1-\sum_{-1}1) &=& \frac{1}{2}\frac{\sum_{+1}r_i^{(t)}(\sum_{+1}1-\sum_{-1}1)}{\sum_{+1}1}-\frac{1}{2}\frac{\sum_{-1}r_i^{(t)}(\sum_{+1}1-\sum_{-1}1)}{\sum_{-1}1}\\
\Rightarrow \hat{w} &=& \frac{1}{2}\left(\frac{\sum_{+1}r_i^{(t)}}{\sum_{+1}1}-\frac{\sum_{-1}r_i^{(t)}}{\sum_{-1}1}\right)
\end{eqnarray*}\qed
\end{proof}



\begin{proof}[\autoref{prop:BGN_dec}]
Let us use the shortcut notation $\sum_{\pm1} := \sum_{i:\sgn(\mathbf{w}_{1,t}\cdot\xbf_i+b_{1,t})=\pm1} $. Note that 
$$
b_{2,t}-w_{2,t} = \frac{\sum_{-1} r_i^{(t)}}{\sum_{-1} 1}\ ,~~~~
b_{2,t}-w_{2,t} = \frac{\sum_{+1} r_i^{(t)}}{\sum_{+1} 1}\ .
$$
We first show that $\ell_S(BGN_{t-1}) - \ell_S(BGN_{t}) > 0, \ \forall t \in \mathbb{N}^*$.
\allowdisplaybreaks[4]
\begin{eqnarray*}
&&\ \quad \ell_S(BGN_{t-1}) - \ell_S(BGN_{t})\\
&&= \frac{1}{m}\sum_{i=1}^{m}\left(r_i^{(t)}-BGN_{t-1}(\mathbf{x}_i)\right)^2 - \frac{1}{m}\sum_{i=1}^{m}\left(r_i^{(t)}-BGN_{t}(\mathbf{x}_i)\right)^2\\
&&= \frac{1}{m}\sum_{i=1}^{m}\left(r^{(t)}_i\right)^2 - \frac{1}{m}\sum_{i=1}^{m}\left(r^{(t)}_i-w_{2,t}h_t(\mathbf{x}_i)-b_{2,t}\right)^2\\
&&= \frac{1}{m}\sum_{i=1}^{m}\left(r^{(t)}_i\right)^2 - \frac{1}{m}\sum_{i=1}^{m}\left(\left(r^{(t)}_i\right)^2-2r^{(t)}_i(w_{2,t}h_t(\mathbf{x}_i)+b_{2,t})+(w_{2,t}h_t(\mathbf{x}_i)+b_{2,t})^2\right)\\
&&= \frac{1}{m}\sum_{i=1}^{m}\left(2r^{(t)}_i(w_{2,t}h_t(\mathbf{x}_i)+b_{2,t})-(w_{2,t}h_t(\mathbf{x}_i)+b_{2,t})^2\right)\\
&&= \textcolor{red}{w_{2,t}\frac{2}{m}\sum_{i=1}^{m}r^{(t)}_ih_t(\mathbf{x}_i)+b_{2,t}\frac{2}{m}\sum_{i=1}^{m}r^{(t)}_i}\textcolor{blue}{-\frac{1}{m}\sum_{i=1}^{m}w_{2,t}^2\underset{=1\forall\mathbf{x}}{\underbrace{h_t^2(\mathbf{x}_i)}}-\frac{1}{m}\sum_{i=1}^{m}b_{2,t}^2}\textcolor{brown}{-w_{2,t}b_{2,t}\frac{2}{m}\sum_{i=1}^{m}h_t(\mathbf{x}_i)}\\
&&= \frac{1}{m}\Bigg[\textcolor{red}{\left(\frac{\sum_{+1}r_i^{(t)}}{\sum_{+1}1}-\frac{\sum_{-1}r_i^{(t)}}{\sum_{-1}1}\right)\left(\sum_{+1}r_i-\sum_{-1}r_i\right)+\left(\frac{\sum_{+1}r_i^{(t)}}{\sum_{+1}1}+\frac{\sum_{-1}r_i^{(t)}}{\sum_{-1}1}\right)\left(\sum_{+1}r_i+\sum_{-1}r_i\right)}\\
&&\ \quad\textcolor{blue}{-\frac{1}{4}\left(\frac{\sum_{+1}r_i^{(t)}}{\sum_{+1}1}-\frac{\sum_{-1}r_i^{(t)}}{\sum_{-1}1}\right)^2\left(\sum_{+1}1+\sum_{-1}1\right)-\frac{1}{4}\left(\frac{\sum_{+1}r_i^{(t)}}{\sum_{+1}1}+\frac{\sum_{-1}r_i^{(t)}}{\sum_{-1}1}\right)^2\left(\sum_{+1}1+\sum_{-1}1\right)}\\
&&\ \quad\textcolor{brown}{-\frac{1}{2}\left(\frac{\sum_{+1}r_i^{(t)}}{\sum_{+1}1}-\frac{\sum_{-1}r_i^{(t)}}{\sum_{-1}1}\right)\left(\frac{\sum_{+1}r_i^{(t)}}{\sum_{+1}1}+\frac{\sum_{-1}r_i^{(t)}}{\sum_{-1}1}\right)\left(\sum_{+1}1-\sum_{-1}1\right)}\Bigg]\\
&&= \frac{1}{m}\Bigg[\textcolor{red}{2\frac{\left(\sum_{+1}r_i^{(t)}\right)^2}{\sum_{+1}1}+2\frac{\left(\sum_{-1}r_i^{(t)}\right)^2}{\sum_{-1}1}}\textcolor{blue}{-\frac{1}{2}\left(\frac{\left(\sum_{+1}r_i^{(t)}\right)^2}{\left(\sum_{+1}1\right)^2}+\frac{\left(\sum_{-1}r_i^{(t)}\right)^2}{\left(\sum_{-1}1\right)^2}\right)\left(\sum_{+1}1+\sum_{-1}1\right)}\\
&&\ \quad\textcolor{brown}{-\frac{1}{2}\left(\frac{\left(\sum_{+1}r_i^{(t)}\right)^2}{\left(\sum_{+1}1\right)^2}-\frac{\left(\sum_{-1}r_i^{(t)}\right)^2}{\left(\sum_{-1}1\right)^2}\right)\left(\sum_{+1}1-\sum_{-1}1\right)}\Bigg]\\
&&= \frac{1}{m}\Bigg[\frac{\left(\sum_{+1}r_i^{(t)}\right)^2}{\sum_{+1}1}+\frac{\left(\sum_{-1}r_i^{(t)}\right)^2}{\sum_{-1}1}\Bigg]>0\\
\end{eqnarray*}
Since we have that 
$$
\left(\sum_{+1}r_i^{(t)}\right)^2\left(\sum_{-1}1\right)+\left(\sum_{-1}r_i^{(t)}\right)^2\left(\sum_{+1}1\right)+\left(\sum_{+1}r_i^{(t)}\right)\left(\sum_{-1}r_i^{(t)}\right)\left(\sum_{+1}1+\sum_{-1}1\right)=0:
$$
$$
\begin{aligned}
\ell_S(BGN_{t-1}) - \ell_S(BGN_{t}) &= \frac{1}{m}\Bigg[\frac{\left(\sum_{+1}r_i^{(t)}\right)^2}{\sum_{+1}1}+\frac{\left(\sum_{-1}r_i^{(t)}\right)^2}{\sum_{-1}1}\Bigg]\\
&= \frac{1}{m}\Bigg[\frac{\left(\sum_{+1}r_i^{(t)}\right)^2\left(\sum_{-1}1\right)+\left(\sum_{-1}r_i^{(t)}\right)^2\left(\sum_{-1}1\right)}{\left(\sum_{+1}1\right)\left(\sum_{-1}1\right)}\Bigg]\\
&= -\frac{1}{m}\Bigg[\frac{m\left(\sum_{+1}r_i^{(t)}\right)\left(\sum_{-1}r_i^{(t)}\right)}{\left(\sum_{+1}1\right)\left(\sum_{-1}1\right)}\Bigg]\\
&= -\left(\frac{\sum_{+1}r_i^{(t)}}{\sum_{+1}1}\right)\left(\frac{\sum_{-1}r_i^{(t)}}{\sum_{-1}1}\right)\\
&= (w_{2,t}+b_{2,t})(w_{2,t}-b_{2,t})\,.
\end{aligned}
$$ 
\qed
\end{proof}